\pdfoutput=1
\documentclass[11pt]{article}

\usepackage{acl}

\usepackage{times}
\usepackage{latexsym}

\usepackage[T1]{fontenc}
\usepackage[utf8]{inputenc}

\usepackage{microtype}

\usepackage[switch]{lineno}
\usepackage{xspace}
\usepackage{colortbl}

\usepackage{booktabs}  
\usepackage{threeparttable}  
\usepackage{multicol}  
\usepackage{multirow}  
\usepackage{mathrsfs}
\usepackage{stfloats}

\usepackage{graphicx} % DO NOT CHANGE THIS
\usepackage{amsmath}
\usepackage{amssymb}
\usepackage{array}
\usepackage{todonotes}
\usepackage{subcaption}
\usepackage{makecell}
\newcolumntype{L}[1]{>{\raggedright\arraybackslash}p{#1}}

\usepackage{arydshln}
\usepackage{pifont}
\newcommand{\cmark}{\ding{51}}
\newcommand{\xmark}{\ding{55}}

\newcommand{\sqlcorrect}[1]{{\color{olive}{\cmark}}}
\newcommand{\sqlwrong}[1]{{\color{red}{\xmark}}}
\usepackage{colortbl}

\definecolor{rdbrdb}{RGB}{205,92,92}

\title{\textit{Before Generation, Align it!} A Novel and Effective Strategy for Mitigating Hallucinations in Text-to-SQL Generation}

\author{Ge Qu $^{1}$, Jinyang Li $^{1}$, Bowen Li $^{2}$, Bowen Qin $^{3}$, Nan Huo$^{1}$, \\ \textbf{Chenhao Ma} $^{4}$, \textbf{Reynold Cheng} $^{1}$ \\
$^{1}$The University of Hong Kong,
$^{2}$ Shanghai AI Laboratory\\
$^{3}$ BAAI,
$^{4}$The Chinese University of Hong Kong, Shenzhen  \\
\texttt{\{quge,jl0725\}@connect.hku.hk}, \texttt{ckcheng@cs.hku.hk} 
}

\begin{document}
\maketitle
\begin{abstract}
Large Language Models (LLMs) driven by In-Context Learning (ICL) have significantly improved the performance of text-to-SQL. Previous methods generally employ a two-stage reasoning framework, namely 1) schema linking and 2) logical synthesis, making the framework not only effective but also interpretable. Despite these advancements, the inherent bad nature of the generalization of LLMs often results in hallucinations, which limits the full potential of LLMs. In this work, we first identify and categorize the common types of hallucinations at each stage in text-to-SQL. We then introduce a novel strategy, Task Alignment (TA), designed to mitigate hallucinations at each stage. TA encourages LLMs to take advantage of experiences from similar tasks rather than starting the tasks from scratch. This can help LLMs reduce the burden of generalization, thereby mitigating hallucinations effectively. We further propose TA-SQL, a text-to-SQL framework based on this strategy. The experimental results and comprehensive analysis demonstrate the effectiveness and robustness of our framework. Specifically, it enhances the performance of the GPT-4 baseline by 21.23\% relatively on BIRD dev and it yields significant improvements across six models and four mainstream, complex text-to-SQL benchmarks. For reproducibility, we release our code and prompt at \url{https://github.com/quge2023/TA-SQL}.

\end{abstract}

\section{Introduction}
\label{sec:intro}
In the age of big data, relational databases, as carriers for storing massive amounts of data, play a crucial role in information processing and data analysis. Text-to-SQL, which aims to convert natural language (NL) queries to executable SQL queries, facilitates access to ubiquitous relational data for a broader range of non-technical users, thereby attracting remarkable attention ~\citep{ijcai2018-553, yu-etal-2018-typesql, wang-etal-2020-rat, lgesql}. 

Recently, Large Language Models (LLMs) have shown impressive success on a wide range of NLP tasks through in-context learning (ICL) \citep{icl}, such as question answering ~\citep{qa1,qa2}, logic reasoning ~\citep{lr1, lr2}, and code generation ~\citep{cg1,cg2}. The application of LLMs has also improved the performance of text-to-SQL to another level of intelligence ~\citep{c3, dinsql,dailsql}. 
Delving into their crafted designs, these works generally approach text-to-SQL through a two-stage paradigm. The first stage, \textbf{Schema Linking}, involves the precise mapping of natural language queries to the relevant entities within a database schema \citep{sl_role,proton,awake}. This meticulous alignment is crucial for the following execution of the query and provides transparency by illustrating how natural language queries are interpreted in relation to the database schema. The second step is \textbf{Logical Synthesis}, which refers to the process of generating accurate SQL queries based on the understanding of the logic of the natural language query and the structure of the database \citep{logical_synthesis}.

Nevertheless, hallucination, a notorious problem in LLMs that refers to the generation of content that is irrelevant, erroneous, or inconsistent with user intents \citep{hallucination_survey}, remains a considerable barrier to current frameworks as a reliable automatic text-to-SQL parser. In this work, we first study and conclude primary hallucinations presented in the aforementioned two stages of current text-to-SQL frameworks and attribute them to two main categories: \textbf{schema-based hallucinations} and \textbf{logic-based hallucinations}, as shown in Table \ref{tab:hallucinations}. Schema-based hallucinations refer to hallucinations in which LLMs might inaccurately identify schema structures, introduce unnecessary attributes, or fail to accurately represent or interpret database values. On the other hand, logic-based hallucinations can also prevent LLMs from executing accurate \texttt{JOIN} operations, applying appropriate SQL clauses such as \texttt{GROUP BY} and nested sub-queries, or computationally reasoning in data science queries. 

\begin{table*}[!t]
\centering
\footnotesize
\begin{tabular}{L{2.4cm}p{3.5cm}p{9cm}}
\toprule
{\bf Schema-Based} & {\bf Definition} &{\bf Example}\\
 \midrule 
Schema Contradiction
\newline (30\%) & 
Refers to the instance where incorrect SQL contradicts schema structure. &
\textbf{Question:} \textit{What language is the set of 180 cards that belongs to the Ravnica block translated into?}
\newline
\textbf{Gold:} SELECT \textbf{{\color{teal}T2.language}} FROM sets AS T1 INNER JOIN \textbf{\color{teal}set\_translations AS T2} ON WHERE T1.block = `Ravnica' AND T1.baseSetSize = 180 
\newline
\textbf{Wrong SQL:} SELECT \textbf{\color{purple}language FROM sets} WHERE baseSetSize = 180 AND block = `Ravnica'\\
\midrule

Attribute Overanalysis
\newline (49\%) & 
Refers to the instance where unnecessary attributes are introduced, leading to a
contradiction with the intended result format.&
\textbf{Question:} \textit{Which player is the tallest?}
\newline
\textbf{Gold:} SELECT \textbf{\color{teal}player\_name} FROM Player ORDER BY height DESC LIMIT 1 
\newline
\textbf{Wrong SQL:} SELECT player\_name, \textbf{\color{purple}height} FROM Player ORDER BY height DESC LIMIT 1\\
\midrule

Value Misrepresentation  
\newline (24\%) & 
Refers to the instance where the model imagines a reasonable but non-existent
value format in the schema.&
\textbf{Question:} \textit{Give the race of the blue-haired men superhero.}
\newline
\textbf{Gold:} SELECT ...  WHERE colour.colour = \textbf{\color{teal}`Blue'} AND gender.gender = \textbf{\color{teal}`Male'}
\newline
\textbf{Wrong SQL:} SELECT ...  WHERE colour.colour = \textbf{\color{purple}`blue'} AND gender.gender = \textbf{\color{purple}`M'}\\

\toprule
{\bf Logic-Based} & {\bf Definition} &{\bf Example}\\

\midrule 
Join Redundancy
\newline (15\%) & 
Refers to the instance where the SQL joins unnecessary tables for complex
text-to-SQL cases.&
\textbf{Question:} \textit{Determine the bond type formed in the chemical compound containing element Tellurium.}
\newline
\textbf{Gold:} SELECT T2.bond\_type \textbf{\color{teal}FROM atom AS T1 INNER JOIN bond} AS T2 ON  WHERE T1.element = `te'
\newline
\textbf{Wrong SQL:} SELECT bond\_type \textbf{\color{purple}FROM bond INNER JOIN connected ON ... INNER JOIN atom} ON ... WHERE atom.element = `te'\\

\midrule 
Clause Abuse
\newline (25\%) & 
Refers to the instance where clauses such as \texttt{GROUP BY} are abused, disrupting
the correct order or limitation of results. &
\textbf{Question:} \textit{Among the posts that were voted by user 14, what is the id of the most valuable post?}
\newline
\textbf{Gold:} SELECT post.Id ...  WHERE votes.UserId = 14 ORDER BY post.FavoriteCount DESC LIMIT 1
\newline
\textbf{Wrong SQL:} SELECT post.Id FROM votes INNER JOIN posts ON ... WHERE votes.UserId = 14 \textbf{\color{purple} GROUP BY post.Id} ORDER BY post.FavoriteCount DESC LIMIT 1\\

\midrule 
Mathematical Delusion 
\newline (17\%) & 
Refers to the instance where the model fails to convert mathematical knowledge
or logic into correct SQL functions, resorting to expressions such as imagined
functions. &
\textbf{Question:} \textit{What is the percentage of the amount 50 received by the Student Club among members?}
\newline
\textbf{Gold:} SELECT CAST(SUM(CASE WHEN income.amount = 50 THEN 1.0 ELSE 0 END) AS REAL) * 100 \textbf{\color{teal}/} COUNT(income.income\_id) FROM ... WHERE member.position = `Member' 
\newline
\textbf{Wrong SQL:} SELECT \textbf{\color{purple}DIVIDE(SUM(CASE WHEN income.amount = 50 THEN 1 ELSE 0 END), COUNT(member.member\_id))} FROM ... WHERE member.position = `Member' \\

 \bottomrule
\end{tabular}
\caption{Definitions and Examples of schema-based and logic-based hallucinations.}
\label{tab:hallucinations}
\end{table*}

The aforementioned challenges reinforce the demand for a robust text-to-SQL framework
to minimize hallucinations and improve overall performance while maintaining interpretability. We posit that hallucinations often arise when models misinterpret the decomposed stages of a task as entirely new challenges, for which they lack prior training. This situation is comparable to human experiences, where unfamiliarity with a task can lead to disorientation and a higher propensity for errors \citep{sql_intro}. Thus, just as experienced individuals can draw on familiar situations to reduce cognitive load and enhance task performance \citep{analogy}, we introduce \textbf{T}ask \textbf{A}lignment (\textbf{TA}), a novel strategy to mitigate hallucinations of LLMs in this way. TA fundamentally adjusts the approach of models to unfamiliar tasks by aligning them with tasks they have previously trained on. This method reduces the dependence of models on their generalization capability for generating responses from scratch, thereby significantly reducing the incidence of hallucinations. 

\begin{figure*}[t]
    \centering
    \includegraphics[width=1.0\textwidth]{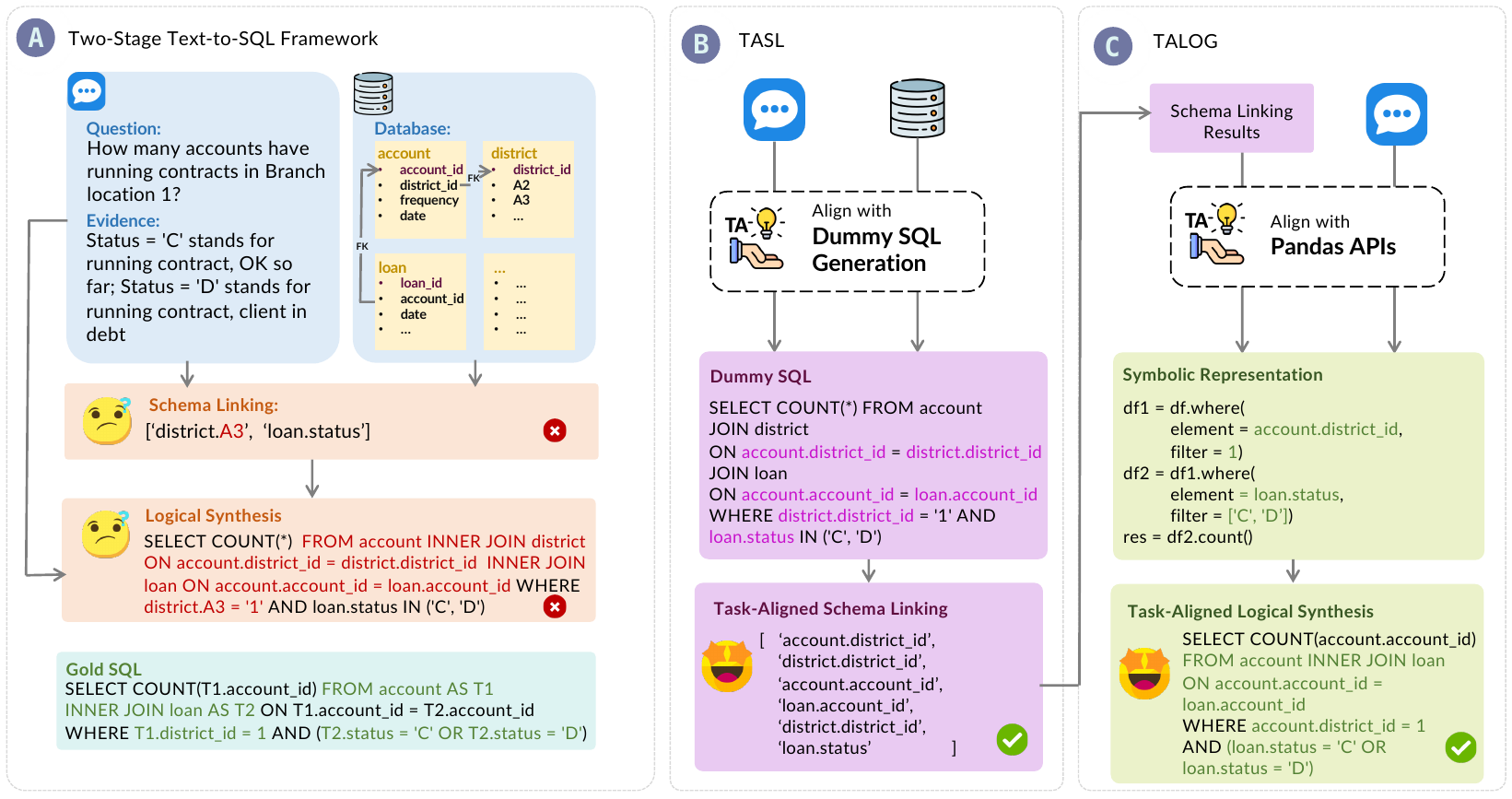}
    \caption{An illustration of TA-SQL, utilizing the TASL (b) and TALOG modules (c), mitigates hallucinations that occur in each of the two stages of previous text-to-SQL frameworks (a).}
    \label{fig:main_figure}
\end{figure*}

We further propose a text-to-SQL framework named \textbf{TA-SQL}, which consists of a \textbf{T}ask-\textbf{A}ligned \textbf{S}chema \textbf{L}inking (\textbf{TASL}) module and a \textbf{T}ask-\textbf{A}ligned \textbf{LOG}ical synthesis (\textbf{TALOG}) module. TA is employed to mitigate hallucinations in these two modules, respectively, enhancing the performance of the framework while preserving its interpretability. Experimental results on four text-to-SQL datasets and our comprehensive analysis demonstrate the effectiveness and robustness of TA-SQL. Specifically, TA-SQL relatively improves the performance of the GPT4 baseline in terms of Execution Accuracy (EX) by 21.23\% and 14.86\% on \textsc{BIRD} \citep{bird} and \textsc{Spider} \cite{spider}, respectively. Moreover, our experimental results also illustrate that TA-SQL is a model-agnostic framework, exhibiting applicability to both mainstream closed-source LLMs and open-source weaker LLMs.

% \patrick{[put in limitation?]Though we only focus on text-to-SQL in this work, we believe that the general TA strategy can be employed to other generative tasks, e.g., Code Retrieval, KBQA, and LLM-Agent Reasoning.} 

\section{Preliminaries} \label{preliminary}
\paragraph{Problem Definition} 
Given a natural language question $\mathcal{Q} = \left\{q_1, ..., q_{\left| \mathcal{Q} \right|}\right\}$ with its corresponding database schema $\mathcal{D} = \left \langle \mathcal{C}, \mathcal{T} \right \rangle$, where $\mathcal{C} = \left\{c_1, ..., c_{\left| \mathcal{C} \right|}\right\}$ and $\mathcal{T} = \left\{t_1, ..., t_{\left| \mathcal{T} \right|}\right\}$ represent columns and tables, $\left| \mathcal{C} \right|$ and $\left| \mathcal{T} \right|$ refer to the number of columns and tables in each database respectively.
The goal of text-to-SQL is to generate the corresponding SQL query $y$.

\paragraph{In-Context Learning}
In-Context Learning (ICL) is a paradigm that allows language models to learn tasks with only a few examples in the form of demonstrations \citep{icl}, or even without examples. It requires no additional training and is directly applicable to pre-trained LLMs. In this work, we only discuss hallucinations in ICL-based text-to-SQL frameworks. Few-shot prompting is a scenario in ICL where it uses task descriptions \(I\) and a set few-shot input-output (I/O) prompting demonstrations \(S = \{(x_1,y_1), ...,(x_k, y_k)\}\) to assist LLM $\mathcal{M}$ to solve the input problem \(x\) which belongs to a task \(m\) by:

\begin{equation}
y = f_{\mathcal{M}}(x,I,S \mid m),
\end{equation}
where $f_{\mathcal{M}}\left(\cdot \mid m \right)$ refers to a mapping function applied by LLM $\mathcal{M}$ when it generalizes task $m$ from scratch. When I/O pairs are no longer provided, the scenario transitions to zero-shot prompting, where the model is expected to understand and complete the task relying solely on its pre-trained knowledge, and the output of zero-shot prompting could be represented as:

\begin{equation}
y = f_{\mathcal{M}}(x,I\mid m),
\label{eq2}
\end{equation}

\section{Methodology}
\subsection{Task Alignment}
\label{subsec: ta}
Inspired by the human approach of drawing upon relevant past experience when tracking unfamiliar tasks, we introduce Task Alignment (TA), a novel strategy designed to mitigate hallucinations. The fundamental idea is that LLMs have already acquired knowledge of various tasks during training \citep{chatgpt}. We refer to tasks for which the basic rules have been mastered by LLMs as pre-trained tasks. Given a novel task \(m^n\) and one of its problems \(x\), TA first retrieves the most related pre-trained task $m^p$ from a set of LLM pre-trained tasks $\{m^p_1, m^p_2, ..., m^p_k\}$. In this study, we manually select $m^p$ for each new task. The potential for LLMs to automatically select tasks is a valuable prospect for future research. It then leverages this pre-trained task to reconstruct the representation for the novel task $m^n$, aligning it with the representation that the LLMs are familiar with. The goal of TA is to solve the problem $x$ by:
\begin{equation}
y = f_{\mathcal{M}}(x,I,S \mid m^n \to m^p),
\end{equation}
where $f_{\mathcal{M}}\left(\cdot \mid m^n \to m^p \right)$ refers to the mapping function applied by LLM $\mathcal{M}$ when it applies experiences from aligned pretrained task $m^p$ while generalizing $m^n$.

TA explicitly guides LLMs to approach unfamiliar tasks from the perspective of more familiar ones, alleviating the burden of from-scratch generalization and subsequently mitigating hallucinations.

\subsection{TA-SQL}
 We further propose a robust text-to-SQL framework named TA-SQL. TA-SQL adheres to the two-stage paradigm of previous work, consisting of a task-aligned schema linking module and a task-aligned logic synthesis module. However, unlike previous works ~\citep {c3, dinsql, dailsql} that treat each decomposed module as an entirely new task for the LLM to generalize from scratch, we apply the TA strategy to each module to stimulate incremental generalization of LLMs. This design not only mitigates hallucinations effectively for better performance but also maintains the interpretability of the entire model. We introduce these two modules, respectively, in the following sections. The prompts employed within each module are displayed in Appendix \ref{sec:module_recipe}.

\subsubsection{Task-Aligned Schema Linking Module (TASL)}
Given a natural language question \(\mathcal{Q}\) with corresponding database schema \(\mathcal{D}\), schema linking is responsible for identifying references to columns, tables, and condition values in \(\mathcal{Q}\). However, LLMs are not adept at the schema linking task. Therefore, when dealing with complex databases characterized by their extensive size and abundant semantic information, LLMs are highly prone to generating schema-based hallucinations. \citep{sl_large}. Hallucinations at this stage would be influential negatively on the final performance due to the error propagation \citep{ep}.

As shown in Figure \ref{fig:main_figure} (b), we design a TASL module for the schema linking stage.
The schema linking task in this module is represented as first generating a dummy SQL query and then extracting related schema entities from it as the final output. Although the schema linking task is not familiar to LLMs, its downstream task, SQL generation, has been extensively exposed during training \citep{guo2024deepseekcoder}. Playing a similar role to negative sampling in the skip-gram algorithm \citep{skip_gram}, the objective of dummy SQL generation is not to create executable SQL directly for the final application. Instead, its primary function is to subtly leverage the successful experiences of schema entity selection during the generation process for LLMs.

\subsubsection{Task-aligned Logical Synthesis Module (TALOG)} \label{sec:reasoning}
The TALOG module is responsible for reasoning the transformation logic from the NL query into SQL based on the results generated by the TASL module and accordingly producing accurate SQL. This process often involves multiple forms of logic, including SQL syntax reasoning, external knowledge reasoning, and computational reasoning. Such complexity presents a significant challenge for LLMs, leading to the emergence of logic-based hallucinations \citep{math-hall}. 

In fact, SQL serves as a tool for extracting values from Relational Database (RDB) for data analysis. It encapsulates various data analysis logics, such as data filtering, mathematical computation, and output synthesis. As such, we employ the TA in the capacity of a data scientist who addresses complex problems through step-by-step logical operations using pandas-like APIs \citep{pandasapi} and generates symbolic representations that include reasoning processes, as shown in Figure \ref{fig:main_figure} (c). 

After logic alignment with data analysis processes, the remaining challenge is to ensure that LLMs are proficient in perceiving valid SQL syntax and structures. This proficiency is crucial for the generation of accurate SQL. To facilitate this, we replace conventional pandas API functions with symbolic functions that resemble SQL keywords, thereby enabling the symbolic representation to invoke them effectively.

\section{Experiments} \label{exp}
\subsection{Experiment Settings}
\paragraph{Datasets.}
We evaluate our text-to-SQL framework on four challenging benchmarks for cross-domain SQLs.  {(1) \textbf {\textsc{Bird}} \citep{bird} is the most challenging lager-scale cross-domain text-to-SQL benchmark. It has two settings, with and without external knowledge, to highlight the new challenges brought by external knowledge. In this paper, we use its development set for evaluation, which contains 1534 pairs of text-to-SQL data and 11 databases, as the test set is not released. \textbf{(2) \textsc{Spider}}\citep{spider} is a more standard cross-domain text-to-SQL benchmark. It contains 1034 examples, which cover 20 complex databases across multiple domains, in the development set. \textbf{(3) \textsc{DK}} \citep{dk} requires text-to-SQL parsers to equip with the capability of domain knowledge reasoning. \textbf{(4) \textsc{Realistic}} removes and switches the obvious mentions of schema items in questions, making it closer to the real scenarios.

\paragraph{Metrics.}
Following the prior study ~\citep{spider, bird}, we use Execution Accuracy (EX) to measure the performance of our method. EX can reflect whether a predicted SQL is valid and return the exact result as the execution result of the ground truth SQL.

\paragraph{Models.}
We experiment our proposed method with both closed-sourced LLMs and open-sourced code generation models. For the closed-source LLMs, we experiment with GPT family models including ChatGPT (\texttt{gpt-3.5-turbo}) \citep{chatgpt}, GPT4 (\texttt{gpt-4-32k}) \citep{openai2023gpt4}, GPT4-Turbo (\texttt{gpt-4-turbo}), and Claude (\texttt{claude-2.0}) \citep{claude2}. For open-source weaker LLM models, we experiment with two most popular and strong baselines, CodeLlama (\texttt{codellama-34b-instruct}) \citep{rozière2023code}, and DeepSeek (\texttt{deepseek-coder-33b-instruct}) \citep{guo2024deepseekcoder}. 

\paragraph{Compared Methods.}
We also compare our method with two SOTA ICL-based methods, that are, DIN-SQL \citep{dinsql} and DAIL-SQL \citep{dailsql} on both \textsc{BIRD} and \textsc{SPIDER}.

\paragraph{Implementation}
We implement the schema linking module with zero-shot prompts and the logical synthesis module with 6-shot prompts. For all models we used in this paper, we set the argument temperature and top-p as 0 and 1, respectively, to promise reproduction. The max\_tokens (\texttt{max\_new\_tokens}) for closed-source LLMs and open-source weaker LLMs are both set as 800, respectively, for all modules. 

\begin{table}[t]
    \centering
    \resizebox{0.9\hsize}{!}{
    \begin{tabular}{lcc}  
    \toprule
    \textbf{\textsc{Method}}& \textbf{\textsc{dev}} & \textbf{\textsc{test}}  \\ 
    \midrule
    \rowcolor{black!10!}\multicolumn{3}{c}{\textbf{\textit{w/o knowledge}}} \\
    Palm-2 & 18.77 & 24.71 \\
    Codex &25.42 &  24.86\\
    ChatGPT & 24.05 & 26.77 \\
    ChatGPT+COT &25.88 & 28.95\\
    Claude-2 & 28.29 & 34.60 \\
    \midrule
    GPT-4     & 30.90 & 34.88 \\
    \rowcolor{blue!15!} TA-SQL+GPT-4 & $\textbf{50.58}_\textbf{ ($\uparrow$ 19.68)}$  & $\textbf{54.38}_\textbf{ ($\uparrow$ 19.50)}$ \\
    \midrule
    \rowcolor{black!10!}\multicolumn{3}{c}{\textbf{\textit{w/ knowledge}}} \\
    Palm-2 &27.38 &33.04 \\
    Codex &34.35 &36.47 \\
    ChatGPT &37.22 & 39.30 \\
    ChatGPT+COT &36.64 & 40.08\\
    Claude-2 & 42.70 & 49.02 \\
    DIN-SQL+GPT-4 $^\clubsuit$ & 50.72 & 55.90 \\
    DAIL-SQL+GPT-4 $^\clubsuit$    &54.76 & 56.08 \\
    \midrule
    GPT-4     & 46.35 & 54.89 \\
    \rowcolor{blue!15!} TA-SQL+GPT-4 & $\textbf{56.19}_\textbf{ ($\uparrow$ 9.84)}$  & $\textbf{59.14}_\textbf{ ($\uparrow$ 4.25)}$ \\

    \bottomrule
    \end{tabular}}
    \caption{Execution Accuracy ({EX}) (\%) on \textsc{BIRD}. $^\clubsuit$ means the model uses self-consistency or re-modification mechanisms. $\uparrow$ is an absolute improvement.}
    \label{tab:main_result}
\end{table}

\subsection{Main Results}
\paragraph{Results on \textsc{BIRD}.} 
Table \ref{tab:main_result} displays the performance of TA-SQL and other competitive methods on the current most challenging text-to-SQL benchmark, \textsc{BIRD}. First, in the setting with oracle knowledge, we demonstrate that TA-SQL effectively mitigates hallucinations in the GPT4 baseline, resulting in a relative improvement of 21.23\% in EX on the development set and 7.74\% on the test set. This demonstrates that even the most powerful LLMs can produce severe hallucinations during the text-to-SQL process, thereby highlighting the value of hallucination mitigation research. \textbf{\textit{Surprisingly, TA-SQL equipped with GPT4 outperforms the SOTA LLM-based method without fine-tuning by 2.61\%}} even without the application of self-consistency or re-modification mechanisms. Furthermore, even in the setting without external knowledge, TA-SQL achieves performance comparable to the GPT4 baseline equipped with oracle external knowledge. This suggests that addressing hallucinations within the existing knowledge could be a promising and cost-effective solution, rather than resorting to the addition of manually extracted external knowledge from heterogeneous resources with much more effort.

\begin{table*}[t]
\centering
\resizebox{1\hsize}{!}{
\begin{tabular}{l|ccccc|ccccc|ccccc}
\toprule
{\multirow{2}*{\textbf{\textsc{Method}}}} & \multicolumn{5}{c|}{\textbf{\textsc{spider}}}      & \multicolumn{5}{c|}{\textbf{\textsc{dk}}}      &  \multicolumn{5}{c}{\textbf{\textsc{realistic}}}   \\
        & easy & medium & hard & extra & all & easy & medium & hard & extra & all & easy & medium & hard & extra & all \\
\midrule
GPT4 & 89.1& 79.8& 61.5&48.8&74.0      &78.2&72.4&50.0&45.7&65.2        &86.2&82.7&57.6&55.7&73.4  \\
\rowcolor{blue!15!} 
+ TA-SQL&\textbf{93.5}&\textbf{90.8}&\textbf{77.6}&\textbf{64.5}&\textbf{85.0}      &\textbf{84.5}&\textbf{78.0}&\textbf{64.9} &\textbf{54.3}&\textbf{72.9}    &\textbf{88.1}&\textbf{87.7}&\textbf{72.7}&\textbf{59.8}&\textbf{79.5}  \\
\bottomrule
\end{tabular}}
\caption{Execution Accuracy ({EX}) across queries of varying levels of difficulty on \textsc{Spider}, \textsc{DK}, and \textsc{Realistic}.}
\label{tab:spider_res}
\end{table*}

\paragraph{Results on \textsc{Spider} and its Variant Datasets} 
As shown in Table \ref{tab:spider_res}, TA-SQL effectively enhances the EX performance of the GPT4 baseline by 14. 86\%, 11. 80\%, and 8. 31\% on the \textsc{Spider} and its variant datasets, \textsc{DK} and \textsc{Realistic}, respectively, with improvements across all difficulty levels. This suggests that TA-SQL, as a general method, is not only useful in complex text-to-SQL scenarios that closely mirror the real world but also delivers robust performance on standard text-to-SQL benchmarks where the context is relatively simple.

\paragraph{Results on Model Agnosticism.} TA-SQL is proved to be model-agnostic since it can work among mainstream closed-source LLM and open-source weaker language models, as shown in Table \ref{tab:model_ad}. TA-SQL can improve the performance across queries of varying difficulty levels for closed-source LLMs. However, we observe that the performance gains brought by TA-SQL for weaker models (CodeLlama, DeepSeek) are relatively limited. This can be attributed to their constrained capabilities in generalizing and instruction-following \citep{generalization}, which limit the effectiveness of TA-SQL in diminishing hallucinations for challenging queries \citep{small_llm}. 

\begin{table}[t]  
    \centering
    \resizebox{0.9\hsize}{!}{
    \begin{tabular}{lcccc}  
    \toprule
    \textbf{\textsc{Model}}& \textbf{\textsc{sim.  }} & \textbf{\textsc{mod.  }} & \textbf{\textsc{chall.}} & \textbf{\textsc{total}} \\ 
    \midrule
    \rowcolor{black!10!}\multicolumn{5}{c}{\textbf{\textit{Closed-Source LLM}}} \\
    \midrule
    GPT4 &54.35&34.64&31.70&46.35\\
    +TA-SQL  &{\cellcolor{blue!20!}63.14}& {\cellcolor{blue!35!}48.60}& {\cellcolor{blue!20!}36.11}& {\cellcolor{blue!20!}56.19}\\
    \midrule
    GPT4-turbo &59.35&38.92&27.78&50.19\\
    +TA-SQL  &{\cellcolor{blue!8!}60.54}& {\cellcolor{blue!8!}40.86}& {\cellcolor{blue!35!}38.19}& {\cellcolor{blue!8!}52.48}\\
    \midrule
    Claude &51.34&30.07&23.24&42.47\\
    +TA-SQL  &{\cellcolor{blue!20!}56.97}& {\cellcolor{blue!20!}39.78}& {\cellcolor{blue!8!}27.78}& {\cellcolor{blue!20!}48.89}\\
    \midrule
    ChatGPT &47.60&22.44&18.31&37.22\\
    +TA-SQL  &{\cellcolor{blue!8!}51.57}& {\cellcolor{blue!20!}33.76}& {\cellcolor{blue!20!}25.69}& {\cellcolor{blue!20!}43.74}\\
    \midrule
    \rowcolor{black!10!}\multicolumn{5}{c}{\textbf{\textit{Open-Source weaker LLM}}} \\
    \midrule
    DeepSeek &51.68&29.03&18.06&41.66\\
    +TA-SQL  &{\cellcolor{blue!8!}53.41}& {\cellcolor{blue!8!}32.04}&
    {\cellcolor{blue!8!}19.44}& {\cellcolor{blue!8!}43.74}\\
    \midrule
    CodeLlama &34.81&15.48&11.11&26.73\\
    +TA-SQL  &{\cellcolor{blue!8!}37.30}& {\cellcolor{pink!25!}13.33}& 11.11&{\cellcolor{blue!8!}27.57} \\
    \bottomrule
    \end{tabular}}
    \caption{Execution Accuracy ({EX}) of TA-SQL employing various models as the backend. \textsc{sim.}, \textsc{mod.}, \textsc{chall.} represent the levels of query difficulty and are the abbreviations of simple, moderate, and challenging, respectively.}
    \label{tab:model_ad}
\end{table}

\subsection{Imperative of Two-stage Paradigm}

\begin{table}[t]  
    \centering
    \resizebox{1.0\hsize}{!}{
    \begin{tabular}{lcccc}  
    \toprule
    \textbf{\textsc{Method}}& \textbf{\textsc{sim.}} & \textbf{\textsc{mod.}} & \textbf{\textsc{chall.}} & \textbf{\textsc{total}} \\ 
    \midrule
    TA-SQL &63.14&48.60&36.11& \textbf{56.19}\\
    w/o Schema Linking  &58.35& 37.92& 32.04& $\textbf{49.77}_\textbf{ ($\downarrow$ 6.42)}$\\
    w/o Logical Synthesis &61.59 & 39.57 & 32.64 & $\textbf{52.41}_\textbf{ ($\downarrow$ 3.78)}$ \\
    \bottomrule
    \end{tabular}}
    \caption{Imperative analysis for the two-stage paradigm on \textsc{BIRD} development set. $\downarrow$ is an absolute decrease.}
    \label{tab:two_stage}
    % \vspace{-0.5}
\end{table}

We conduct an imperative analysis of the two-stage paradigm. Table \ref{tab:two_stage} illustrates that the two-stage paradigm not only makes the text-to-SQL framework interpretable, but also impacts the overall performance of the framework. Specifically: 
\paragraph{The schema linking module constitutes a prerequisite for the success of TA-SQL.} 
Firstly, the removal of the schema linking module disrupts the interpretability of the framework,  preventing it from correcting hallucinations through more flexible methods such as human-computer interaction. Secondly, through quantitative analysis, we discover that the removal of the schema linking module leads to a significant performance decline across queries of varying difficulty levels ($\downarrow$ 6.42\% in total). This is attributed to the fact that more accurate schema linking results not only reduce schema-based hallucinations but also facilitate the subsequent logical synthesis module to conduct more granular and complex reasoning based on these results.

\paragraph{The logical synthesis module determines the upper bound for the performance of the entire framework.} This is evidenced by the observation that, relative to the performance decline on simple queries ($\downarrow$ 1.55\%), the removal of the logical synthesis module has a more obvious impact on moderate ($\downarrow$ 10.68\%) and challenging queries ($\downarrow$ 3.74\%). Relatively challenging queries often contain more complex analytical intentions, involving mathematical computations, multi-step reasoning, or compositional generalization. The symbolic representation produced by the logical synthesis module effectively guides analytical reasoning processes, thereby raising the upper bound of the framework's ability to solve complex problems.

\subsection{Ablation Study}
\label{sec:ablation}
After validating the imperative of the two-stage paradigm, we further conduct an ablation study to evaluate the effectiveness of implementing the TA strategy within these two stages. The schema linking module and logical synthesis module, following the customized designs in DIN-SQL \citep{dinsql} and NatSQL\citep{natsql}, are implemented, respectively, for comparison.

\begin{figure}
    \centering
    \includegraphics[width=0.4\textwidth]{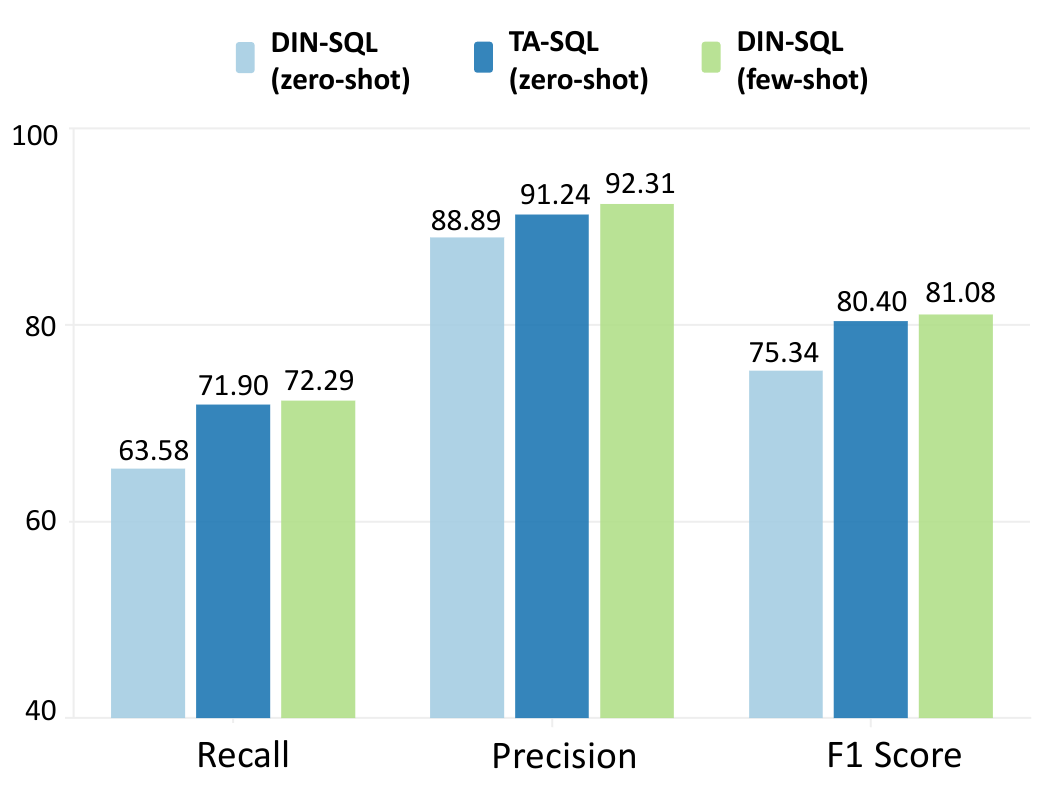}
    \caption{Results of different schema linking modules on BIRD-dev.}
    \label{fig:ta_sl}
    % \vspace{-0.5cm}
\end{figure}

\paragraph{Results in the Schema Linking Stage.} We implement the schema linking module of DIN-SQL in both the zero-shot and few-shot settings for comparison. Three metrics are introduced to facilitate a more intuitive comparison of the schema linking results: (1) \textbf{Recall} computes the ratio of instances in which the schema linking outcomes encompass all ground truth schema elements of this instance. (2) \textbf{Precision} quantifies the accuracy of the linked schema. (3) \textbf{F1 Score} represents a harmonic mean of recall and precision. The detailed definitions of these metrics are presented in Appendix \ref{sec:sl_metric}.

\begin{figure}
    \centering
    \includegraphics[width=0.4\textwidth]{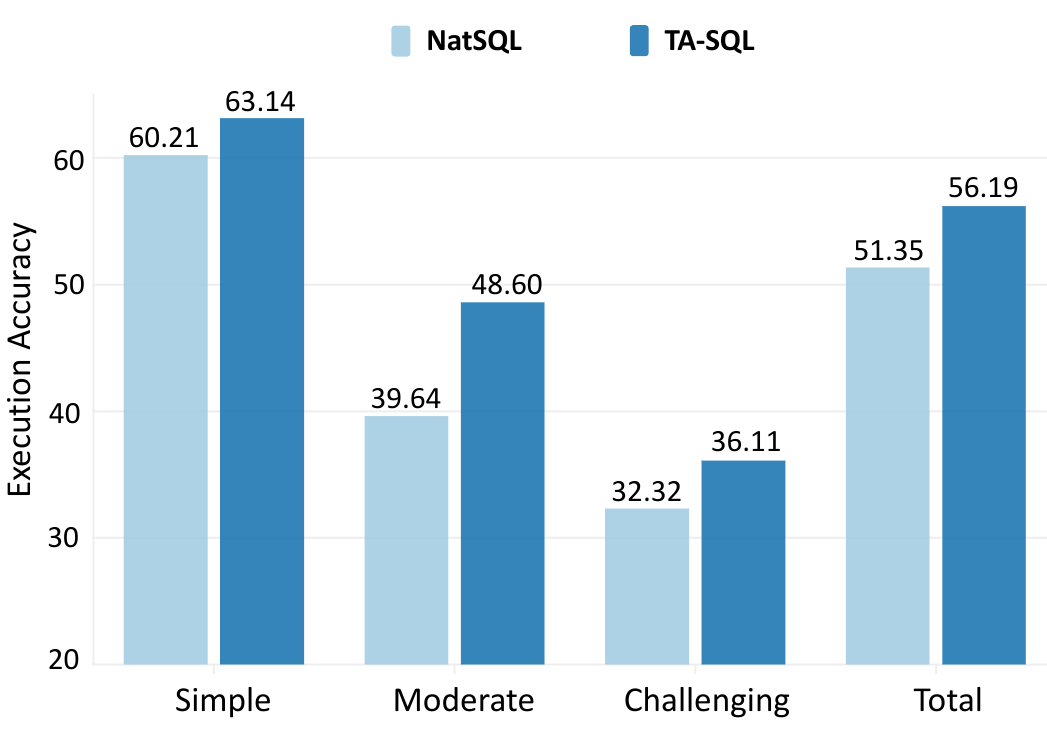}
    \caption{Results of different logical synthesis modules on BIRD-dev.}
    \label{fig:ta_r}
    % \vspace{-0.5cm}
\end{figure}

Figure \ref{fig:ta_sl} illustrates that, even when following the design of the SOTA method DIN-SQL, zero-shot schema linking tasks confuse LLMs, since it requires LLMs to comprehend and generalize this unfamiliar task from scratch. This confusion can be alleviated by human-annotated example demonstrations in the few-shot setting (F1 Score: 75.34 $\to$ 81.08). However, the TASL module in the zero-shot setting can directly achieve results that are competitive with the few-shot schema linking module in DIN-SQL without human intervention. This suggests that TA can effectively guide the model to align with pre-trained knowledge to tackle unfamiliar tasks without relying on additional external information.

\paragraph{Results in the Logical Synthesis Stage.} We implement another logical synthesis module with a classic customized symbolic representation designed in NatSQL \citep{natsql} and refer to it as NatLOG module. Same as TALOG, it is also implemented in the 6-shot setting. As shown in Figure \ref{fig:ta_r}, the TALOG module exhibits superior performance
across various levels of difficulty compared with the NatLOG module. This is because the custom symbolic representation includes new rules that the model needs to understand and learn from scratch, thereby increasing the emergence of hallucinations during the logical synthesis process.

\subsection{Fine-grained Case Study} \label{sec:case_study}

% \begin{figure}
%     \centering
%     \includegraphics[width=0.45\textwidth]{case_study.pdf}
%     \caption{Case study: two illustrative cases from \textsc{BIRD} development set. Ex1 is from california\_school database. And Ex2 is from debit\_card\_specializing database.}
%     \label{fig:case_study}
% \end{figure}

\begin{table*}[t]  
\centering
\small
%\resizebox{0.95\hsize}{!}{
\scalebox{0.8}{
\begin{tabular}{llc}
\toprule
\rowcolor{blue!15!} \textbf{\textcolor{black}{Case 1}} & \texttt{california\_school}& \\ 
 \midrule
 \leavevmode {\bf \color{orange}Question} & \textit{Which active district has the highest average score in Reading?} &  \\
 \leavevmode {\bf \color{blue}Gold} & SELECT {\color{blue}T1.District}  FROM  {\color{blue}schools AS T1} INNER JOIN satscores  AS  T2 ON T1.CDSCode =  T2.cds WHERE  & \\
 & T1.StatusType = 'Active'  ORDER BY  {\color{blue}T2.AvgScrRead}  DESC  LIMIT  1
 &  \\ 

 \hdashline
  \leavevmode {\bf \color{purple}GPT4} & SELECT {\color{purple}dname, AVG(AvgScrRead)} FROM satscores JOIN schools ON satscores.cds = schools.CDSCode & \sqlwrong{} \\ 
 & WHERE StatusType = 'Active'  {\color{purple}GROUP  BY  dname}  ORDER BY {\color{purple} AVG(AvgScrRead)} DESC LIMIT 1 &  \\
 \hdashline
 \leavevmode {\bf \color{teal}TA-SQL} & \texttt{df1 = df.where(element = schools.StatusType, filter = 'Active')} &  \\
 & \texttt{df2 = df1.orderby(by = {\color{teal}satscores.AvgScrRead}, desc).limit(1)}
&\\
&\texttt{res = df2.select({\color{teal}schools.District})}
&\\ 
 & SELECT {\color{teal}schools.District} FROM satscores INNER JOIN schools ON satscores.cds = schools.CDSCode WHERE 
 & \sqlcorrect{} \\
  & schools.StatusType = 'Active' ORDER BY {\color{teal}satscores.AvgScrRead} DESC LIMIT 1 &  \\

 \midrule
 \rowcolor{blue!15!}\textbf{\textcolor{black}{Case 2 
}} &\texttt{debit\_card\_specializing} & \\ 
  \midrule
 \leavevmode {\bf \color{orange}Question} & \textit{What is the percentage of the customers who used EUR in 2012/8/25?} &  \\
 \leavevmode {\bf \color{blue}Gold} & SELECT CAST(SUM(IIF(T2.Currency = 'EUR', 1, 0))  AS FLOAT)  * 100  /  {\color{blue}COUNT( T1.CustomerID )}  & \\
 & FROM transactions\_1k AS T1 INNER JOIN customers AS T2 ON T1.CustomerID = T2.CustomerID 
 &  \\ 
 & {\color{blue}WHERE T1.Date  =  '2012-08-25'}
 &  \\ 

 \hdashline
  \leavevmode {\bf \color{purple}GPT4} & SELECT (SELECT COUNT(CustomerID) * 100.0 / {\color{purple}(SELECT COUNT(CustomerID) FROM customers)}  & \sqlwrong{} \\ 
 & FROM transactions\_1k AS T1 INNER JOIN customers AS T2 ON T1.CustomerID = T2.CustomerID &  \\
 &WHERE customers.Currency = 'EUR' AND transactions\_1k.Date = '2012-08-25')&\\
 \hdashline
 \leavevmode {\bf \color{teal}TA-SQL} & \texttt{df1 = df.where(element = transactions\_1k.Date, {\color{teal}filter = '2012-08-25')}} &  \\
 & \texttt{df2 = df1.where(element = customers.Currency, filter = 'EUR')}
&\\
&\texttt{res = df.select(cast(df2.count(), real) * 100 / {\color{teal}df1.count()})}
&\\ 
  & SELECT CAST(COUNT(*) * 100.0 / {\color{teal}(SELECT COUNT(*) FROM transactions\_1k WHERE Date =}  
 & \sqlcorrect{} \\
  & {\color{teal}'2012-08-25') AS REAL)} FROM transactions\_1k AS T1 INNER JOIN customers AS T2 ON T1.CustomerID  &  \\
  & = T2.CustomerID WHERE transactions\_1k.Date = '2012-08-25' AND customers.Currency = 'EUR'
&\\
 \bottomrule
\end{tabular}}
%}
\caption{Case study: two illustrative cases from BIRD-dev.}
\label{tab:case}
% \vspace{-0.5}
\end{table*}

To illustrate the effectiveness of TA-SQL qualitatively, we conduct an analysis of its performance across databases in the \textsc{BIRD} development set. TA-SQL demonstrates a significant impact on databases \texttt{debit\_card\_specializing}, \texttt{toxicology}, and \texttt{california\_schools}, yielding relative EX increases of 59.05\%, 43.1\%, and 32.13\%, respectively. \textbf{\textit{From the perspective of database schema}}, as shown in Table \ref{tab:case} Case 1, GPT4 selects \texttt{satscores.dname} and \texttt{AVG(AvgScrRead)} as final attributes of the generated SQL, which contradicts the intent of the NL query. This suggests that GPT4 struggles to map NL query entities to the database schema within complex contexts, leading to the occurrence of schema contraction and attribute overanalysis, which fall under schema-based hallucinations. 
However, TA-SQL can correspond the question entities to the correct column names through precise retrieval of tables and columns by the TASL module, coupled with more granular schema-related reasoning, such as element selection, by the TALOG module. \textbf{\textit{From the perspective of query difficulty}}, the capacity of TA-SQL to mitigate logic-based hallucinations can yield a more substantial effect within databases that contain complex queries, which require multiple logical operations. In Case 2, GPT4 manifested erroneous computational logic, which is an instance of mathematical delusion, suggesting GPT4's limited capability when confronted with complicated multi-step reasoning. Conversely, TA-SQL clearly demonstrates the data manipulation process, thereby equipping it with the capacity to manage complex logic.

\subsection{Discussion about Hallucinations in Text-to-SQL Systems}
Given that this is the first attempt to systematically study hallucinations in text-to-SQL systems, we define them following the survey \citep{ha_nlg}. All types of hallucinations shown in Table \ref{tab:hallucinations} are categorized as \textbf{Intrinsic Hallucinations}, which occur when the generated SQL query contradicts the information or intent expressed in the natural language query, the underlying database schema, or SQL syntax. 

It is worth noting that the distinction between hallucinations we define and errors is quite subtle. While hallucinations can lead to the occurrence of errors, they may not always directly result in errors. For instance, joining redundant tables can sometimes produce the same executed results, which are considered correct in the current SQL evaluation system. More importantly, errors are typically detected after the final SQLs are outputted and executed, usually when the entire workflow is completed. However, hallucinations can be observed and mitigated before final result generations (i.e., during schema linking or logical synthesis phases). Our proposed method originates from this problem definition and achieves equivalent or better performance than error-corrected methods, as demonstrated in the experiment section. Figure \ref{fig:hall_dist} shows the fine-grained performance of TA-SQL on mitigating hallucinations across each category.

\begin{figure*}[ht]
    \centering
    \includegraphics[width=1.0\textwidth]{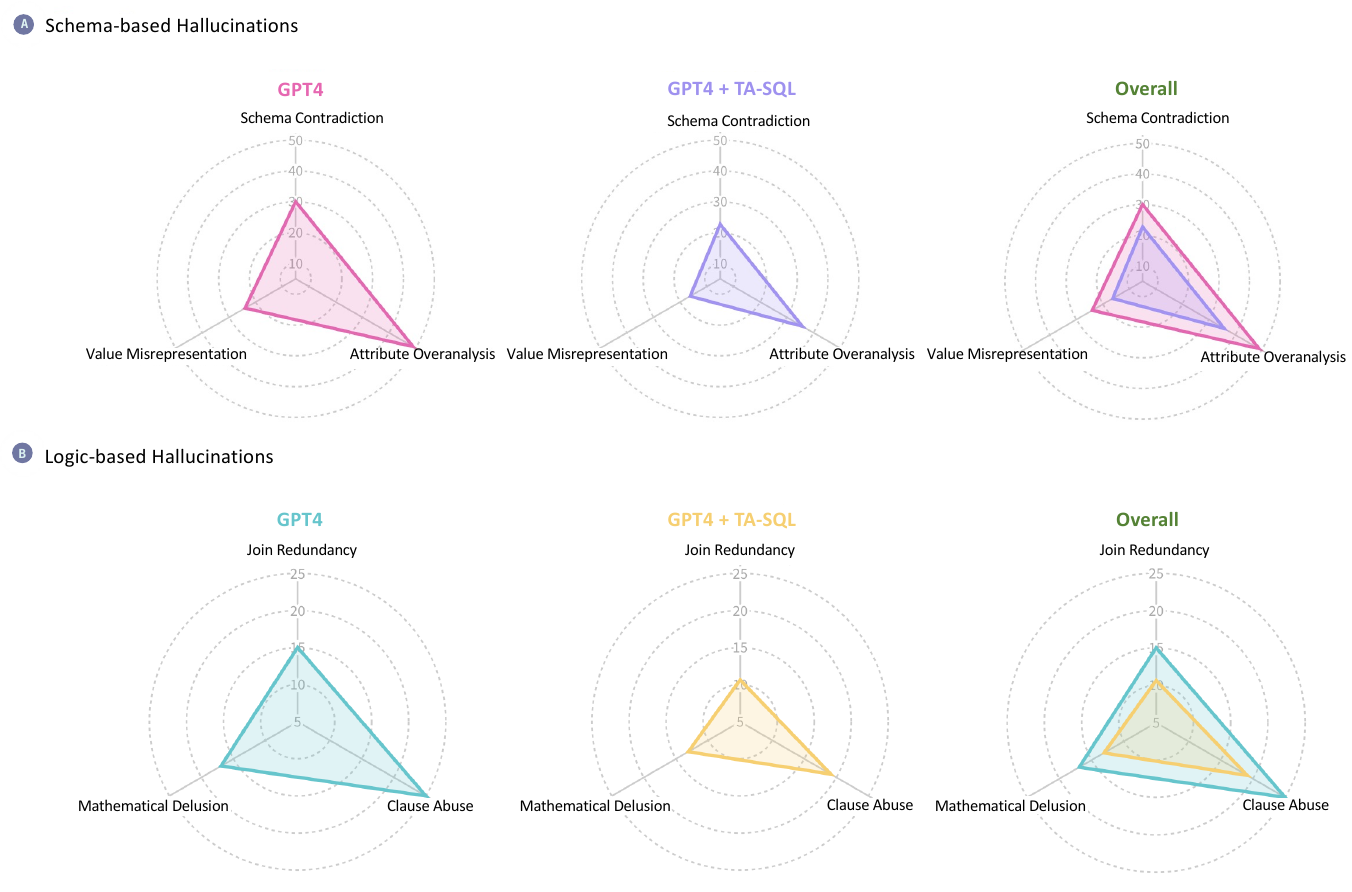}
    \caption{Performance of fine-grained categorical hallucination mitigation on BIRD.}
    \label{fig:hall_dist}
\end{figure*}

\section{Related Work}
\paragraph{Text-to-SQL}

The development of a successful cross-domain text-to-SQL parser fundamentally involves the creation of an encoder for learning representations of questions and schema and a decoder for generating SQL queries \citep{t2s_survey}. For instance, RATSQL \citep{wang-etal-2020-rat}, SDSQL \citep{sdsql}, LGESQL \citep{lgesql}, S$^2$SQL \citep{s2sql}, and Proton \citep{proton} have advanced the representation learning of natural language questions and database schema using a relational graph neural network. The introduction of sequence-to-sequence pre-trained language models (PLMs) such as T5 \citep{t5} and NQG-T5 \citep{nqg-t5} significantly transforms text-to-SQL tasks, given their adaptability and generative capabilities across diverse datasets. These models demonstrate impressive results through fine-tuning with minimal effort. Besides, PICARD \citep{picard} designs a constrained decoder to reject inadmissible tokens at the decoding step. RASAT \citep{rasat} further improves the structural information encoding of T5 by integrating schema alignment into the encoder, while Graphix \citep{graphix} has equipped T5 with multi-hop reasoning. RESDSQL \citep{resdsql} enhances T5 by decoupling the schema linking and the skeleton parsing. 

Recently, large language models (LLMs) \citep{chatgpt, palm, claude2} have attracted considerable attention due to their robust reasoning and domain generalization capabilities. Models like DIN-SQL \citep{dinsql} and DAIL-SQL \citep{dailsql} with few-shot demonstrations, along with the evolution of language models to language agents ~\citep{mind2web, middleware}, have pioneered text-to-SQL solutions to a new level of intelligence. 

\paragraph{Hallucination}
One of the most prominent challenges is hallucination, a phenomenon where a model generates information that is not present or inferred from the input \citep{ha_nlg}. This issue is particularly serious in text generation tasks, as evidenced by the factual consistency problems of dialog generation \citep{dialog_ha_1, dialog_ha_2, dialog_ha_3} when using LLMs. As one of the important techniques in database applications, hallucination can result in the text-to-SQL generation of erroneous or non-sensical SQL queries. 

\section{Conclusion}
In this research, we first systematically identify and classify common hallucination types in text-to-SQL. Subsequently, we propose Task Alignment (TA), a novel strategy to mitigate hallucinations in Large Language Models (LLMs) during the text-to-SQL process. Based on this strategy, we further propose TA-SQL, a framework to mitigate hallucinations at each stage of this process. Experimental results and comprehensive analysis show the importance of hallucination research in text-to-SQL and data science and suggest promising directions for future work.

\section{Limitations}
Our findings suggest that TA is particularly adept at handling complex cases where the knowledge evidence supplied by the BIRD database is explicit and the questions poised are unequivocally answerable, as shown in Section \ref{sec:case_study}. On the contrary, in the databases \texttt{codebase\_community}, \texttt{student\_club}, and \texttt{thrombosis\_prediction}, we observe that the clarity of questions and the sufficiency of knowledge evidence render it more effective to directly generate SQL queries from annotated data, bypassing the need for a multi-step calibration process. Furthermore, the selection of familiar tasks in each phase of the text-to-SQL conversion process is currently conducted by human prior knowledge rather than an automated mechanism capable of identifying and retrieving relevant and familiar tasks for TA applications. This gap highlights a notable avenue for future research.

\section{Acknowledgement}
We thank all constructive comments from anonymous reviewers. Reynold Cheng, Ge Qu, Jinyang Li, and Nan Huo were supported by the Hong Kong Jockey Club Charities Trust (Project 260920140), the University of Hong Kong (Project 109000579), and the HKU Outstanding Research Student Supervisor Award 2022-23. Chenhao Ma was partially supported by NSFC under Grant 62302421, Basic and Applied Basic Research Fund in Guangdong Province under Grant 2023A1515011280. Ge Qu and Jinyang Li were supported by HKU Presidential PhD Scholar Programme. Ge Qu was also funded by Hong Kong PhD Fellowship Scheme.

\section{Ethical Statement}
All datasets employed in this work are publicly accessible, ensuring the transparency and reproducibility of our findings. Furthermore, the output generated by our investigations is structured as SQL queries—a programming language format—rather than natural language text, which could potentially involve harmful or biased content. Our team meticulously examines each output to confirm the absence of politically sensitive or biased material. Finally, regarding our analysis of open-source models utilizing GPUs, such as Deepseek and Codellama, it is notable that our approach involves only model inference without training. 
% This methodology can mitigate the environmental impact, specifically in terms of carbon dioxide emissions, associated with computational processes.
\newpage

\bibliography{acl_latex}
\bibliographystyle{acl_natbib}

\clearpage
\appendix
\section{TA-SQL Recipe}
\label{sec:module_recipe}

\subsection{Pythonic Prompt}
Unlike previous works that used natural language to construct prompts, all prompts used in TA-SQL are pythonic prompts. Considering that text-to-SQL is fundamentally a code generation task, pythonic prompts can exhibit data structures more clearly and express constraints and requirements more concisely.

\subsection{TASL Module}
The schema linking task in TASL module is represented as first generating a dummy SQL query and then extracting related schema entities from it as the final output. As discussed in Section \ref{sec:ablation}, we implement the TASL module in the zero-shot setting, leveraging the efficient employment of TA. The zero-shot prompt used to generate dummy SQL in this module is presented in Figure \ref{fig:direct_sql_prompt}. 

\begin{figure*}[ht]
    \centering
    \includegraphics[width=1.0\textwidth]{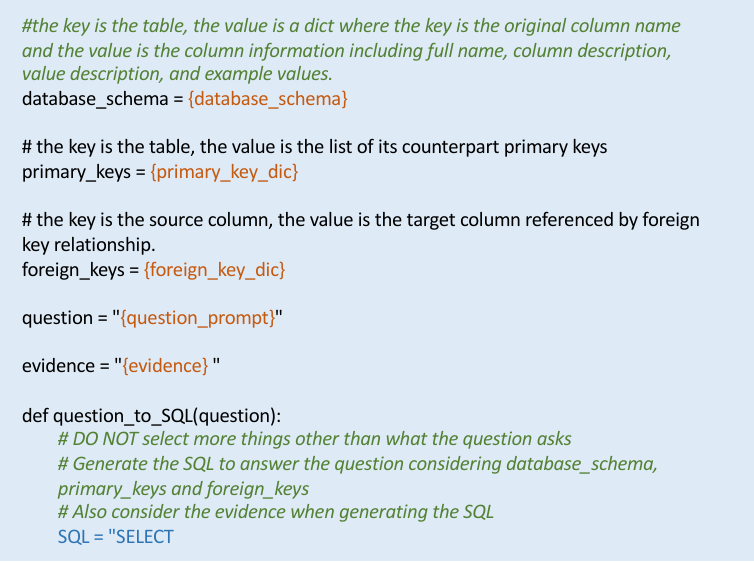}
    \caption{The prompt of generating dummy SQLs.}
    \label{fig:direct_sql_prompt}
\end{figure*}

Specifically, we employ a Python dictionary to represent the database schema, where the key is the \texttt{table\_name.column\_name} entity (e.g. \texttt{account.account\_id}), and the value is the comprehensive description of the corresponding column. However, directly concatenating all related information, such as column type, original column description, and value description, as the final comprehensive description might lead to an excessively lengthy prompt. This verbosity could potentially confuse LLMs. Therefore, to prevent such issues, as a preparatory step to generating dummy SQL, we first prompt LLMs to generate a succinct description for each column, drawing upon the aforementioned related information. These succinct descriptions then serve as the value for each column within the database schema dictionary. The prompt used to generate succinct column descriptions is presented in Figure \ref{fig:cc_prompt}.

\begin{figure*}[ht]
    \centering
    \includegraphics[width=1.0\textwidth]{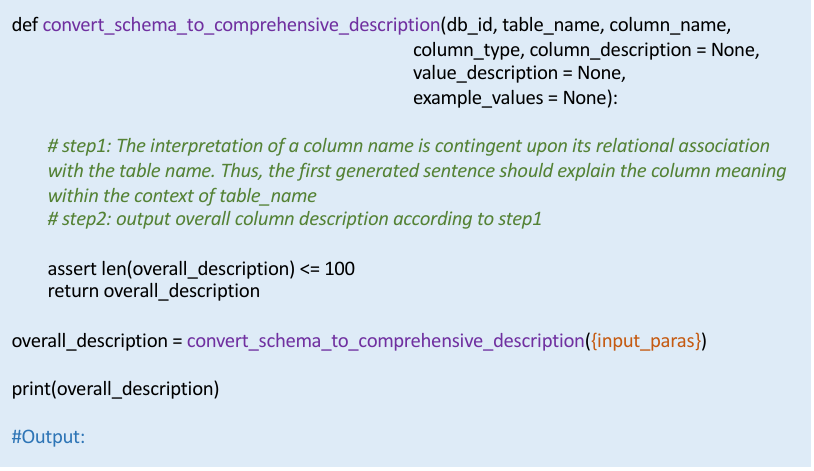}
    \caption{The prompt of succinct column description generation.}
    \label{fig:cc_prompt}
\end{figure*}

\subsection{TALOG Module}
TALOG module employs pandas APIs to guide LLMs in conducting step-by-step logical reasoning. However, there is a natural gap between the conventional pandas APIs and the ultimate execution language, SQL. To bridge this, we replace the conventional pandas API functions with symbolic functions that resemble SQL keywords, thereby ensuring a precise translation from generated symbolic representations to SQLs in subsequent steps. To facilitate the model's understanding of this substitution, we provide a few demonstrations, specifically six shots, for the model to learn from, thereby generating the desired symbolic representations. Figure \ref{fig:few_shot} presents the prompts used within TALOG for generating symbolic representations. 

\begin{figure*}[ht]
    \centering
    \includegraphics[width=1.0\textwidth]{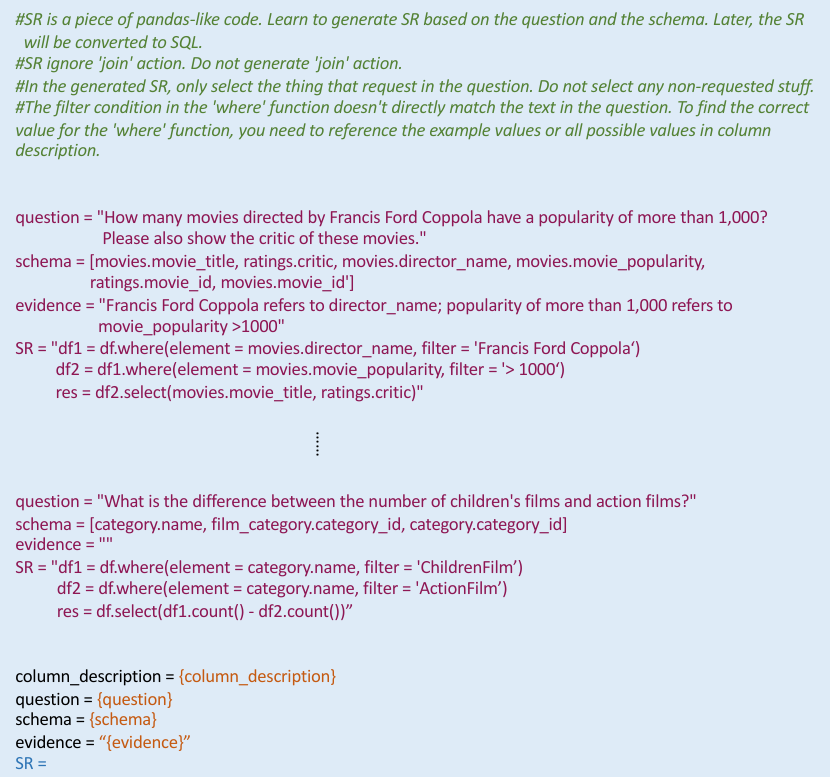}
    \caption{The prompt of generating symbolic representations.}
    \label{fig:few_shot}
\end{figure*}

\section{Details of Evaluation Metrics for Schema Linking}
\label{sec:sl_metric}

\paragraph{Recall.} Recall \(P\) is defined as the proportion of queries for which the linked schema outputted by the schema linking module contains all the ground truth schema, relative to the overall number of queries. It is noteworthy that since the schema retrieved by the TASL module would replace the original complete database schema as the input for the subsequent TALOG modules, the recall determines the upper bound of the EX for the final generated SQL. Considering the ground truth schema set as \(S_n\) of the \(n^{th}\) query, and the linked schema set as \(\hat{S_n}\), EM could be computed by:
\begin{equation}
\small
R = \frac{\sum_{n=1}^N \mathbb{I}(\hat{S}_n, S_n)}{N}
\end{equation}
where \(\mathbb{I}(\hat{S}_n, S_n)\) is an indicator function, which can be represented as
\begin{equation}
\small
\mathbb{I}(\hat{S}, S) = 
\begin{cases}
    1, & \hat{S} \supseteq S \\
    0, & \hat{S} \nsupseteq S
\end{cases}
\label{eq_indicator}
\end{equation} \\ 
% \textbf{(2) F1 Score}: The F1 score is a measure of schema linking result's accuracy, considering both the precision \(P\) and the recall of the \(R\) of the result. 
\paragraph{Precision.} Precision \(P\) quantifies the accuracy of the linked schema. Considering the ground truth schema set as \(S_n\) with length \(L_n\) of the \(n^{th}\) query, and the linked schema set as \(\hat{S_n}\) with length \(\hat{L}_n\), precision is computed by:

\begin{equation}
P = \frac{\sum_{n=1}^{N} p_n}{N}, p_n = \frac{\sum_{j=1}^{\hat{L}_n} \mathbb{I}(\hat{s}_j \in S_n)}{\hat{L}_n}
\label{eq_p}
\end{equation}

\paragraph{F1 score.} F1 score \(F1\) represents a harmonic mean of recall and precision. It offers an evaluation of schema linking results, taking into account both precision and recall as:
\begin{equation}
\small
F1 = \frac{2 \cdot P \cdot R}{P+R}
\end{equation}

% \section{More Results on \textsc{Spider}}

% \begin{table}[t]  
%     \centering
%     \resizebox{0.4\hsize}{!}{
%     \begin{tabular}{lc}  
%     \toprule
%     \textbf{\textsc{Method}}& 
%     \textbf{\textsc{EX}}\\ 
%     \midrule
%     C3 & 81.80\\
%     DIN-SQL &82.80 \\
%     DAIL-SQL  &84.40 \\
%     TA-SQL &85.00\\

%     \bottomrule
%     \end{tabular}}
%     \caption{Execution Accuracy (EX) on Spider dev for TA-SQL in comparison with three SOTA ICL-based methods.}
%     \label{tab:more_spider_res}
%     % \vspace{-0.5}
% \end{table}

% As shown in Table \ref{tab:more_spider_res}, similar to results on \textsc{BIRD}, TA-SQL surpasses all these three ICL-based methods in terms of Execution Accuracy (EX) on Spider dev.

\section{Implement Details}
The open-source models are implemented using \texttt{Pytorch}\footnote{https://pytorch.org/}, \texttt{Transformers}\footnote{https://huggingface.co/docs/transformers/en/installation}, and \texttt{vllm}\footnote{https://github.com/vllm-project/vllm}. To expedite the inference process, we also implemented \texttt{deepspeed}\footnote{https://github.com/microsoft/DeepSpeed}. The DeepSeek and CodeLlama models are accessed via \texttt{huggingface}\footnote{https://huggingface.co/models}.

\end{document}